\documentclass[conference]{IEEEtran}
\usepackage{makecell}
\usepackage[utf8]{inputenc}
\usepackage{graphicx}
\usepackage{ragged2e}
\usepackage{url}

\title{Emotion Talk: Emotional Support via Audio Messages for Psychological Assistance}

\author{
\IEEEauthorblockN{
    \begin{minipage}[t]{0.45\textwidth}
    \centering
    Fabrycio Leite Nakano Almada\\
    \textit{Instituto de Informática} \\
    \textit{Universidade Federal de Goiás}\\
    Goiânia, Brasil \\
    fabrycio@discente.ufg.br
    \end{minipage}
    \hfill
    \begin{minipage}[t]{0.45\textwidth}
    \centering
    Kauan Divino Pouso Mariano\\
    \textit{Instituto de Informática} \\
    \textit{Universidade Federal de Goiás}\\
    Goiânia, Brasil \\
    kauan@discente.ufg.br
    \end{minipage}
    \\
    \\
    \vspace{10mm}
    \begin{minipage}[t]{0.45\textwidth}
    \centering
    Maykon Adriell Dutra\\
    \textit{Instituto de Informática} \\
    \textit{Universidade Federal de Goiás}\\
    Goiânia, Brasil \\
    maykonadriell@discente.ufg.br
    \end{minipage}
    \hfill
    \begin{minipage}[t]{0.45\textwidth}
    \centering
    Victor Emanuel da Silva Monteiro\\
    \textit{Instituto de Informática} \\
    \textit{Universidade Federal de Goiás}\\
    Goiânia, Brasil \\
    victor\_emanuel@discente.ufg.br
    \end{minipage}
}
}

\begin{document}

\maketitle

\begin{abstract}
This paper presents "Emotion Talk," a system designed to provide continuous emotional support through audio messages for psychological assistance. The primary objective is to offer consistent support to patients outside traditional therapy sessions by analyzing audio messages to detect emotions and generate appropriate responses. The solution focuses on Portuguese-speaking users, ensuring that the system is linguistically and culturally relevant. This system aims to complement and enhance the psychological follow-up process conducted by therapists, providing immediate and accessible assistance, especially in emergency situations where rapid response is crucial. Experimental results demonstrate the effectiveness of the proposed system, highlighting its potential in applications of psychological support.
\end{abstract}


\section{Introduction}
The increasing demand for psychological support services necessitates continuous assistance beyond traditional therapy sessions. With the growing awareness and acceptance of mental health issues, more individuals are seeking professional help, which places a significant burden on the available psychological resources. Consequently, there is a pressing need for systems that can provide immediate and accessible psychological support, particularly in emergency situations where response time is critical.

Emotion Talk is designed to fill this gap by offering continuous support to patients through audio messages. This system is particularly valuable for patients who may experience emotional distress outside of regular therapy hours. By leveraging advanced audio processing and natural language processing (NLP) techniques, Emotion Talk can analyze the emotional content of audio messages and provide timely, relevant responses that help stabilize the patient until they can consult their therapist. The solution is specifically designed for Portuguese-speaking users, ensuring cultural and linguistic appropriateness in its responses.

The system's potential applications extend beyond individual therapy sessions. Emotion Talk can be integrated into broader mental health support frameworks, providing a scalable solution for clinics and hospitals. Moreover, it can serve as a valuable tool in remote or underserved areas where access to mental health professionals is limited. By offering a reliable and immediate form of support, Emotion Talk contributes to the overall well-being of individuals, helping them navigate emotional crises effectively.

\section{Methodology}
To develop a robust system that accurately identifies and responds to emotional states through audio messages, it is essential to incorporate advanced techniques in audio processing, transcription, emotion detection, natural language processing (NLP), response generation and email integration. Each component plays a critical role in ensuring the overall effectiveness and reliability of the system. By leveraging state-of-the-art models and methodologies, Emotion Talk is designed to provide real-time psychological assistance, with a focus on linguistic and cultural appropriateness for Portuguese-speaking users. The following sections detail the specific methodologies and technologies employed in the development of each module of the system, starting with audio processing.

\subsection{Audio Processing}
The audio processing module is crucial for the initial handling of the patient's audio messages. It is responsible for loading audio files, adjusting their sampling rate, and converting them into a Mel spectrogram. This standardization is essential for ensuring consistent and accurate analysis in subsequent steps. Techniques such as Short-Time Fourier Transform (STFT) and resampling are employed to achieve data uniformity.

The audio processing starts with loading the audio file using torchaudio, which supports various audio formats and provides efficient loading capabilities. Once loaded, the audio is resampled to a target sampling rate, which is critical for maintaining compatibility with the processing pipeline. The use of a Mel spectrogram helps in emphasizing the perceptually important aspects of the audio signal, making it easier to detect emotional cues.

In addition to resampling and spectrogram conversion, padding or trimming the audio to a fixed length is performed to ensure uniform input dimensions for the emotion detection model. This preprocessing step is vital for maintaining the integrity of the machine learning pipeline. By standardizing the audio input, the system can more accurately extract and analyze emotional features, leading to more reliable emotion detection.

\subsection{Audio Transcription using Whisper}
The Whisper model is utilized for transcribing audio messages into text, enabling further analysis and response generation. This model provides high-quality transcription, ensuring that the emotional content of the messages is accurately captured. The transcribed text is then processed using the NLP module to classify the sentiment and generate appropriate responses. The use of Whisper enhances the system's capability to handle diverse audio inputs and maintain the accuracy of the transcription process.

\subsection{Emotion Detection}
The emotion detection module is a core component of the Emotion Talk system. It employs the EMOTION2VEC+ model, which has been specifically trained to recognize and categorize emotions in audio recordings. The model maps the predicted emotions to labels in English, such as 'angry', 'happy', 'neutral', 'sad', or 'unknown'. This mapping process is crucial for interpreting the results and generating appropriate responses.

The process begins with feature extraction from the preprocessed audio. The Mel spectrogram serves as the input to the emotion detection model, which processes the spectrogram to identify emotional cues. The model's architecture is designed to capture a wide range of emotional nuances, allowing it to provide accurate predictions even with diverse audio inputs.

The detected emotions are then used to inform the response generation module. By accurately identifying the emotional state of the patient, the system can tailor its responses to be empathetic and relevant. This emotion-aware interaction enhances the patient's experience and ensures that the support provided is aligned with their current emotional needs.

\subsection{Natural Language Processing (NLP)}
The NLP module plays a significant role in processing transcribed audio text and generating sentiment classifications. It utilizes a pre-trained BERT model, which is known for its robust performance in various NLP tasks. This model transforms the transcribed texts into sentiment classifications such as "sad", "neutral", and "happy", providing a deeper understanding of the emotional context.

The sentiment analysis process involves feeding the transcribed text into the BERT model, which outputs a sentiment label based on the text's content. This label is then mapped to a corresponding emotion, helping the system understand the patient's emotional state. The use of a pre-trained model ensures that the sentiment analysis is both accurate and efficient.

In addition to sentiment analysis, the NLP module is also responsible for ensuring that the generated responses are contextually appropriate and empathetic. By analyzing the patient's messages, the system can generate responses that are not only emotionally relevant but also supportive and reassuring. This level of understanding and empathy is crucial for providing effective psychological support.

\subsection{Response Generation}
The response generation module is responsible for producing contextual responses based on the detected emotions in the audio and transcribed text. It utilizes the GPT-3.5 Turbo language model, which is renowned for its ability to generate human-like text. The model is prompted with a combination of the patient's emotional state and their transcribed messages to generate a coherent and empathetic response.

The process starts by constructing a prompt that includes the patient's emotional context and message history. This prompt is then fed into the GPT-3.5 Turbo model, which generates a response that aims to address the patient's emotional needs. The response generation is guided by a system prompt that ensures the generated text is empathetic and supportive.

In our experiments, we also tested a version of the Llama 3 8B Instruct model and the Zephyr 7B Beta model with different prompt engineering techniques. However, the GPT-3.5 model followed the instructions and expected response format more accurately than the other models. This superior performance in adhering to the desired response format and content makes GPT-3.5 the preferred choice for our system.

By combining emotion detection and advanced language modeling, the system can provide responses that are both relevant and comforting. This approach helps in stabilizing the patient's emotional state and offers immediate psychological support. The use of a sophisticated language model ensures that the responses are nuanced and contextually appropriate, enhancing the overall effectiveness of the system.

\subsection{Report Generation and Email Integration}
The system includes a module for generating reports that summarize the patient's interactions and emotional states over time. These reports are valuable for psychologists to monitor the patient's progress and tailor their treatment plans accordingly. The report generation process involves compiling the conversation history, analyzing the emotions detected, and summarizing key points.

To facilitate effective communication between the system and psychologists, an email service is integrated to send these reports directly to the assigned psychologists. The email content includes the full conversation history, emotion summaries, and any significant observations. This automated report generation and email delivery ensure that psychologists are kept informed about their patients' emotional well-being in a timely manner.

\section{Experimental Results}
The experimental setup for evaluating the Emotion Talk system involved training a Convolutional Neural Network (CNN) on the emoUERJ dataset, which contains a diverse set of emotional audio recordings. The dataset was split into 80\% training and 20\% testing data to ensure a robust evaluation of the model's performance. During the inference phase with new audio samples, we observed difficulties in accurately classifying emotions, which prompted the testing of alternative models such as Emotion2Vec. Ultimately, the Emotion2Vec+ model emerged as the best-performing model, achieving an accuracy of 0.76 and an F1 score of 0.77.

The emoUERJ dataset was specifically chosen for this study because it comprises audio recordings in Portuguese, aligning with the target language of the Emotion Talk system. As the final solution is designed to support Portuguese-speaking users, utilizing a dataset in the same language ensures that the emotion detection models are trained and tested under realistic conditions. This linguistic consistency enhances the model's ability to accurately interpret and respond to the emotional nuances present in the audio messages of the target user base.

In addition to evaluating the model's performance, the experimental results also demonstrated the system's ability to provide timely and relevant responses. The integration of emotion detection with response generation allowed the system to offer immediate support to patients, which is crucial in emergency situations. The overall effectiveness of the system was validated through various test scenarios, highlighting its potential in real-world applications.

Furthermore, techniques of prompt engineering were utilized to enhance the response quality of the GPT-3.5 Turbo language model. By carefully designing and refining the prompts fed into the model, the system ensures that the generated responses are contextually appropriate and empathetic. This method optimizes the interaction quality, providing users with more accurate and supportive feedback based on their emotional state and message content.

Furthermore, the system's architecture ensures scalability and flexibility, allowing it to be integrated into different mental health support frameworks. The experimental results provide a strong foundation for further development and deployment of the Emotion Talk system. The promising performance of the system indicates its potential to significantly enhance the provision of psychological support.

\begin{table}[!htbp]
\centering
\setlength{\tabcolsep}{4pt}
\small
\begin{tabular}{|l|c|c|}
    \hline
    \textbf{Model} & \textbf{Accuracy} & \textbf{F1 Score} \\
    & \textbf{emoUERJ} & \textbf{emoUERJ} \\
    \hline\hline
    Emotion2Vec+ & 0.76 & 0.77 \\
    \hline
    \makecell[l]{Emotion Recognition \\ Wav2Vec IEMOCAP} & 0.47 & 0.40 \\
    \hline
    \makecell[l]{Wav2Vec Base Speech \\ Emotion Recognition} & 0.43 & 0.36 \\
    \hline
    \makecell[l]{Sentiment Predictor \\ for Stress Detection} & 0.16 & 0.17 \\
    \hline
\end{tabular}
\vspace{2mm}
\caption{Comparison of Model Performance on emoUERJ Dataset}
\label{tab:model_comparison}
\end{table}

The table above shows the performance comparison of different emotion detection models on the emoUERJ dataset. The Emotion2Vec+ model outperformed the other models with an accuracy of 0.76 and an F1 score of 0.77. This demonstrates the model's superior ability to recognize and classify emotions in audio recordings.

The Emotion Recognition Wav2Vec IEMOCAP model, while still relatively effective, achieved lower performance metrics with an accuracy of 0.47 and an F1 score of 0.40. Similarly, the Wav2Vec Base Speech Emotion Recognition model showed moderate performance with an accuracy of 0.43 and an F1 score of 0.36.

The Sentiment Predictor for Stress Detection model had the lowest performance among the compared models, with an accuracy of 0.16 and an F1 score of 0.17. This indicates that it is less suitable for emotion detection tasks compared to the other models evaluated.

These results highlight the effectiveness of the Emotion2Vec+ model and validate its selection for the Emotion Talk system. Its ability to accurately detect emotions ensures that the responses generated by the system are appropriate and supportive, enhancing the overall effectiveness of psychological assistance provided.

\section{Database Schema}
The database schema for the Emotion Talk system is designed to ensure efficient and organized management of data related to psychologists, patients, and their interactions. The schema includes tables for psychologists, patients, and conversations, each with specific attributes to capture relevant information. This relational database structure supports the seamless integration and retrieval of data, which is essential for the system's functionality.

The psychologists' table includes columns for storing the psychologist's ID, name, and email address. This information is crucial for managing the assignments of patients to their respective psychologists. Each psychologist can have multiple patients, creating a one-to-many relationship between the psychologists and patients tables. This relationship is implemented using foreign keys to ensure data integrity and consistency.

The patients' table includes columns for storing the patient's ID, name, and the ID of their assigned psychologist. This linkage allows for easy retrieval of all patients associated with a particular psychologist. Additionally, each patient can have multiple conversations, which are stored in the conversations table. The conversations table includes columns for storing the conversation ID, patient ID, and the conversation text. This structure supports the chronological storage and retrieval of patient interactions, facilitating effective follow-up and support.

The database schema's design ensures that all interactions are recorded and can be accessed efficiently. This structure supports the system's overall functionality by providing a reliable way to manage and retrieve critical information. The organized storage of data also aids in the generation of reports and summaries for therapists, enhancing their ability to monitor and support their patients effectively.

\section{Discussion}

The implementation and results of the Emotion Talk system indicate a significant advancement in providing continuous psychological support through audio messages. By integrating state-of-the-art technologies in audio processing, emotion detection, and natural language processing, the system offers a scalable and effective solution for mental health support. The high accuracy and F1 score of the Emotion2Vec+ model underscore its reliability in real-world applications, particularly in recognizing and classifying a wide range of emotions from audio inputs.

One of the key strengths of the Emotion Talk system is its ability to deliver immediate and contextually relevant responses. This is particularly crucial in emergency situations where rapid emotional stabilization is necessary. The use of the GPT-3.5 Turbo language model ensures that the generated responses are not only empathetic but also coherent and supportive, enhancing the overall user experience.

The integration of the Whisper model for audio transcription further strengthens the system by providing accurate and high-quality transcriptions of audio messages. This enables the NLP module to perform precise sentiment analysis, thereby improving the overall effectiveness of the response generation process. The combination of these advanced models ensures that the system can handle diverse audio inputs while maintaining high levels of accuracy and relevance in its responses.

Furthermore, the automated report generation and email delivery feature provide psychologists with valuable insights into their patients' emotional states and interaction history. This functionality not only aids in monitoring patient progress but also helps in tailoring treatment plans to address specific emotional needs. The ability to compile and summarize conversation histories into concise reports enhances the overall efficiency and effectiveness of psychological follow-up.

However, there are areas that require further exploration and improvement. While the Emotion2Vec+ model performs well, there is always room for enhancing its accuracy and ability to detect more nuanced emotions. Additionally, the system's reliance on a stable internet connection for accessing the GPT-3.5 Turbo API could be a limitation in remote or underserved areas. Future developments could focus on creating more robust offline capabilities to ensure uninterrupted support.

Overall, the Emotion Talk system represents a promising step forward in the domain of psychological assistance. Its comprehensive approach to emotion detection and response generation, coupled with robust data management and reporting features, positions it as a valuable tool for both patients and mental health professionals. Continuous improvement and adaptation of the system will be essential to meet the evolving needs of psychological support services.

\section{Conclusion}
The Emotion Talk system effectively integrates audio processing, emotion detection, and natural language processing to provide continuous psychological support. The experimental results validate the system's potential in delivering immediate and accessible assistance, making it a valuable tool in psychological support applications. The system's ability to analyze emotional content in audio messages and generate empathetic responses ensures that patients receive timely and relevant support, especially in emergency situations.

It is important to emphasize that the Emotion Talk system is designed to complement, not replace, the role of psychologists. By facilitating continuous support, the system allows patients to record and manage their emotional experiences in real-time, even when they are away from therapy sessions. This capability helps bridge the gap between sessions, providing valuable insights that can enhance the therapeutic process. The tool aids psychologists by offering detailed emotional logs, thereby enabling more informed and efficient treatment planning.

The scalability and flexibility of the Emotion Talk system allow it to be adapted for various mental health support frameworks, including clinics and hospitals. The system's architecture supports the efficient management of psychologist and patient data, ensuring that support is provided in a structured and organized manner. The promising performance of the system highlights its potential to significantly enhance the provision of psychological support and improve patient outcomes.

While the current implementation focuses on Portuguese-speaking users, the methodology and architecture of the system can be expanded to support additional languages, broadening its applicability and impact. Future work will focus on expanding the system's capabilities and improving its accuracy and responsiveness. This includes refining the emotion detection model, enhancing the natural language processing capabilities, and integrating additional features to support a wider range of psychological conditions. By continuing to develop and optimize the Emotion Talk system, we aim to provide an even more robust and effective tool for psychological support that assists and augments the therapeutic work of psychologists.

\end{document}